\PassOptionsToPackage{unicode}{hyperref}
\PassOptionsToPackage{hyphens}{url}
\PassOptionsToPackage{dvipsnames,svgnames,x11names}{xcolor}
\documentclass[
  12pt]{article}

\usepackage{amsmath,amssymb}
\usepackage{caption}
\usepackage{iftex}
\ifPDFTeX
  \usepackage[T1]{fontenc}
  \usepackage[utf8]{inputenc}
  \usepackage{textcomp} 
\else 
  \usepackage{unicode-math}
  \defaultfontfeatures{Scale=MatchLowercase}
  \defaultfontfeatures[\rmfamily]{Ligatures=TeX,Scale=1}
\fi
\usepackage{lmodern}
\ifPDFTeX\else  
\fi
\IfFileExists{upquote.sty}{\usepackage{upquote}}{}
\IfFileExists{microtype.sty}{
  \usepackage[]{microtype}
  \UseMicrotypeSet[protrusion]{basicmath} 
}{}
\makeatletter
\@ifundefined{KOMAClassName}{
  \IfFileExists{parskip.sty}{%
    \usepackage{parskip}
  }{
    \setlength{\parindent}{0pt}
    \setlength{\parskip}{6pt plus 2pt minus 1pt}}
}{
  \KOMAoptions{parskip=half}}
\makeatother
\usepackage{xcolor}
\makeatletter
\ifx\paragraph\undefined\else
  \let\oldparagraph\paragraph
  \renewcommand{\paragraph}{
    \@ifstar
      \xxxParagraphStar
      \xxxParagraphNoStar
  }
  \newcommand{\xxxParagraphStar}[1]{\oldparagraph*{#1}\mbox{}}
  \newcommand{\xxxParagraphNoStar}[1]{\oldparagraph{#1}\mbox{}}
\fi
\ifx\subparagraph\undefined\else
  \let\oldsubparagraph\subparagraph
  \renewcommand{\subparagraph}{
    \@ifstar
      \xxxSubParagraphStar
      \xxxSubParagraphNoStar
  }
  \newcommand{\xxxSubParagraphStar}[1]{\oldsubparagraph*{#1}\mbox{}}
  \newcommand{\xxxSubParagraphNoStar}[1]{\oldsubparagraph{#1}\mbox{}}
\fi
\makeatother

\usepackage{longtable,booktabs,array}
\usepackage{calc} 
\usepackage{multirow}
\usepackage{etoolbox}
\makeatletter
\patchcmd\longtable{\par}{\if@noskipsec\mbox{}\fi\par}{}{}
\makeatother
\IfFileExists{footnotehyper.sty}{\usepackage{footnotehyper}}{\usepackage{footnote}}
\makesavenoteenv{longtable}
\usepackage{graphicx}
\makeatletter
\def\maxwidth{\ifdim\Gin@nat@width>\linewidth\linewidth\else\Gin@nat@width\fi}
\def\maxheight{\ifdim\Gin@nat@height>\textheight\textheight\else\Gin@nat@height\fi}
\makeatother
\setkeys{Gin}{width=\maxwidth,height=\maxheight,keepaspectratio}
\makeatletter
\def\fps@figure{htbp}
\makeatother

\makeatletter
\@ifpackageloaded{caption}{}{\usepackage{caption}}
\AtBeginDocument{%
\ifdefined\contentsname
  \renewcommand*\contentsname{Table of contents}
\else
  \newcommand\contentsname{Table of contents}
\fi
\ifdefined\listfigurename
  \renewcommand*\listfigurename{List of Figures}
\else
  \newcommand\listfigurename{List of Figures}
\fi
\ifdefined\listtablename
  \renewcommand*\listtablename{List of Tables}
\else
  \newcommand\listtablename{List of Tables}
\fi
\ifdefined\figurename
  \renewcommand*\figurename{Figure}
\else
  \newcommand\figurename{Figure}
\fi
\ifdefined\tablename
  \renewcommand*\tablename{Table}
\else
  \newcommand\tablename{Table}
\fi
}
\@ifpackageloaded{float}{}{\usepackage{float}}
\floatstyle{ruled}
\@ifundefined{c@chapter}{\newfloat{codelisting}{h}{lop}}{\newfloat{codelisting}{h}{lop}[chapter]}
\floatname{codelisting}{Listing}

\makeatother
\makeatletter
\@ifpackageloaded{caption}{}{\usepackage{caption}}
\@ifpackageloaded{subcaption}{}{\usepackage{subcaption}}
\makeatother

\ifLuaTeX
  \usepackage{selnolig}  
\fi
\usepackage{authblk}
\usepackage[]{natbib}
\usepackage{bookmark}

\IfFileExists{xurl.sty}{\usepackage{xurl}}{} 
\hypersetup{
  pdftitle={Bounded AUKS: A Novel Evaluation Metric for Anomaly Detection via Bounded Area Under the Kolmogorov--Smirnov Curve},
  pdfauthor={Yanshuo Zhang; Juan Du},
  pdfkeywords={Evaluation metrics, Distribution separability, Threshold selection},
  colorlinks=true,
  linkcolor={blue},
  filecolor={Maroon},
  citecolor={Blue},
  urlcolor={Blue},
  pdfcreator={LaTeX via pandoc}}

\begin{document}

\def\spacingset#1{\renewcommand{\baselinestretch}%
{#1}\small\normalsize} \spacingset{1}


\title{\bf Bounded AUKS: A Novel Evaluation Metric for Anomaly Detection via Bounded Area Under the Kolmogorov--Smirnov Curve}
\author[1]{Yanshuo ZHANG}
\author[1,2,3]{Juan DU\thanks{Email: yzhang420@connect.hkust-gz.edu.cn; juandu@ust.hk (corresponding author).}}
\affil[1]{Smart Manufacturing Thrust, The Hong Kong University of Science and Technology (Guangzhou), Guangzhou, 100190, China}
\affil[2]{Data Science and Analytics Thrust, The Hong Kong University of Science and Technology (Guangzhou), Guangzhou, 100190, China}
\affil[3]{Department of Mechanical and Aerospace Engineering, The Hong Kong University of Science and Technology, Hong Kong SAR, 999077, China}
\date{}
\maketitle
\thispagestyle{empty}

\bigskip
\begin{abstract}
Recent literature widely used global metrics such as AUROC and AUPRO to quantify the detection performance.
AUROC and AUPRO scores are approaching 100\% in some public datasets such as MVTec AD dataset and VisA dataset, which misleadingly suggests that the pixel-level anomaly detection task has already been solved. 
Threshold-independent metrics represented by AUROC inherently ignore the threshold selection process, which is inevitable in practical applications. Although AUPRO restricts the False Positive rates (FPRs) to an acceptable range, extremely low FPRs can lead to increased false negative rates (FNRs). We argue that there is a lack of metrics capable of quantifying the separation between the anomaly score distributions of normal and anomalous samples, which hinders advancements in anomaly detection. 
In response, we propose the \emph{\textbf{Bounded} \textbf{A}rea \textbf{U}nder the \textbf{K}olmogorov--\textbf{S}mirnov curve} (\textbf{Bounded AUKS}), a partial integral metric that integrates the difference between TPR and FPR within a data-driven bounded range.
Specifically, we introduce AUKS from KS integral as a direct measure of separability between the anomaly score distributions.
The bounded range is determined only by score distributions of normal test samples. Lower bound excludes the uninformative high-FPR region. Upper bound avoids evaluation in extreme regions that aggressively push FPR towards zero, which result in a very low recall.
Our experiments on 9 models across MVTec AD and VisA datasets show that Bounded AUKS can address the limitations of existing metrics and has a better performance in evaluating unsupervised anomaly detection models.
\end{abstract}

\noindent%
{\it Keywords:} evaluation metrics, distribution separability, threshold selection
\vfill

\newpage
\spacingset{1.8} 

\section{Introduction}\label{sec1}

Anomaly detection is a critical task in the field of vision-based inspection, fundamentally aiming to accurately distinguish anomalies from normals. This task is based on normal patterns, where models must identify anomalous features or behavioral patterns that deviate from normality~\citep{Yan2017ImageAnomaly,Austin2024EmergentAnomaly}. In visual inspection involving images, videos and 3D point-cloud data, the initial objective is typically to determine whether an anomaly is present at the instance level~\citep{Du2025FEM3DADSurvey}. However, mere instance-level detection is often insufficient. For example, PointSGRADE combines sparse learning with a graph representation to detect surface anomalies in unstructured 3D point clouds~\citep{Tao2025PointSGRADE}. In image-based inspection, anomaly localization requires precisely identifying the spatial extent of anomalies at the pixel level~\citep{Bertoldo2024AUPIMO, Rafiei2023PixelLevel}.

In most modern anomaly localization pipelines, models output a continuous anomaly score $S$ (or a pixel-wise anomaly map), which will be converted into a hard prediction via a decision threshold $\theta$ (e.g., predicting anomaly when $S>\theta$)  to obtain actionable decisions in deployment. Notably, the threshold is not an auxiliary detail: it directly determines the precision--recall trade-off, controls the false positive rate (FPR), and the shape of predicted anomaly regions. In unsupervised settings, anomaly labels are not available to calibrate $\theta$ on a test set, so the selection of practical thresholds is typically based on anomaly-free statistics (i.e., independent of information on anomalies)~\citep{wilkinghoff2023f1ev,Bertoldo2024AUPIMO}. Therefore, the existing evaluation protocol that involves sweeping over thresholds implicitly presumes a plausible set of candidate thresholds.

The rapid development of anomaly detection methods has also raised increasing demands on evaluation metrics. Specifically, after converting continuous anomaly scores into binary predictions through a decision threshold, model performance can be assessed using conventional metrics such as Accuracy and F1 score, as defined in Supplementary Material Section~S2, Table~S2. Meanwhile, anomaly detection studies also commonly adopt threshold-independent metrics to evaluate model performance from a global perspective, including the Area Under the Receiver Operating Characteristic Curve (AUROC)~\citep{Fawcett2006ROC,Bergmann2021MVTecmetric} and the Area Under the Precision-Recall Curve (AUPR)~\citep{davis2006AUPR}.
For the task of anomaly localization, specialized evaluation metrics have been developed to assess region-level detection performance. A prominent example is the Area Under the Per-Region Overlap (AUPRO) curve~\citep{Bergmann2021MVTecmetric}, which emphasizes detection sensitivity at low false positive rates (FPR), a regime of particular importance for practical industrial deployment.
Despite these advances, such evaluation metrics still suffer from three major limitations:

\begin{itemize}
    \item \textbf{Inflated Scores:} The AUROC scores are approaching 100\% on the open dataset MVTec Anomaly Detection (MVTec AD)~\citep{Bergmann2019MVTecdataset,Bergmann2021MVTecmetric} and Vision Anomaly (VisA)~\citep{Zou2022SPotTheDifference}, which misleadingly suggests that the pixel-level anomaly detection task has already been solved~\citep{Bertoldo2024AUPIMO}. This is an inflated figure. 
    \item \textbf{Lack of Separability Measure:} Existing metrics primarily assess ranking quality or detection accuracy, but they do not explicitly quantify the degree of separation between the score distributions of normal and anomalous samples. In other words, they cannot indicate whether the two distributions are well separated with minimal overlap.
    \item \textbf{Disregard for Threshold Selection:} Threshold-dependent metrics such as F1-max capture only the single best operating point, offering no insight into performance stability across thresholds. AUROC bypasses threshold selection entirely, providing no deployment guidance. AUPRO restricts evaluation to a fixed low-FPR regime, a choice that may not suit all applications. 
\end{itemize}

\begin{table}[!ht]
    \centering
    \caption{Recent high-performing anomaly detection models with near-perfect AUROC scores on MVTec AD and VisA benchmarks. Pixel-level scores are shown.}
    \label{tab:intro_auroc_saturation}
    \fontsize{9pt}{11pt}\selectfont
    \setlength{\tabcolsep}{3pt}
    \begin{tabular}{lccc}
        \toprule
        \textbf{Model} & \textbf{First Publication} & \textbf{MVTec AD AUROC} & \textbf{VisA AUROC} \\
        \midrule
        Dinomaly~\citep{Guo2025Dinomaly} & 2025-06 &  98.6 &  99.1 \\
        HETMM~\citep{2025HETMMmodel} & 2025-05 &  99.0 &  99.1 \\
        URD~\citep{liu2024URDmodel} & 2025-02 &  
        99.0 &  99.1 \\
        GLASS~\citep{chen2024GLASSmodel} & 2024-11 &  99.3 &  98.8 \\
        GLAD~\citep{yao2024GLADmodel} & 2024-10 &  98.6 &  98.6 \\
        \bottomrule
    \end{tabular}
\end{table}

Regarding the first limitation, Table~\ref{tab:intro_auroc_saturation} presents the pixel-level AUROC results of representative recent anomaly detection models on the MVTec AD and VisA datasets. As illustrated, many recent state-of-the-art (SOTA) models report near-perfect pixel-level AUROC scores on MVTec AD and VisA benchmarks. However, these high AUROC values often obscure significant shortcomings in anomaly localization performance. 

\begin{figure}[!ht]
    \centering
    \includegraphics[width=0.95\linewidth]{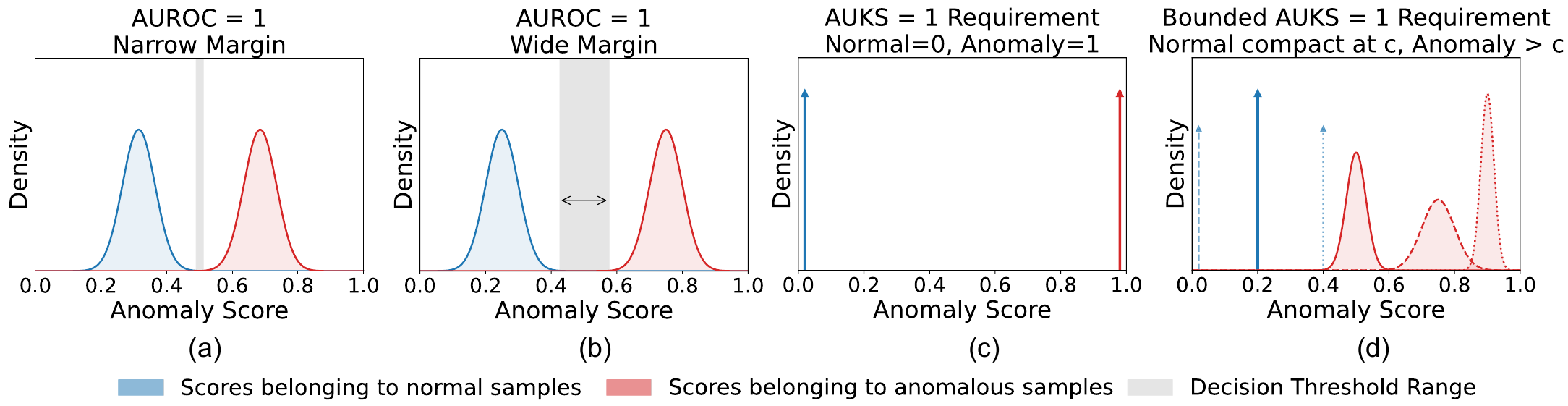}
    \caption{Comparison of ideal score distributions under different metrics. 
    (a) A narrow margin under AUROC = 1.0.
    (b) A wide margin under AUROC = 1.0.
    Both examples in (a) and (b) achieve the same AUROC, but they differ fundamentally in robustness of threshold selection. 
    (c) The strict requirement for AUKS = 1.0, in which normal-scores are 0 and anomalous-scores are 1, underscores that a distinct separation between distributions makes the threshold easier to determine.
    (d) Bounded AUKS = 1.0. A perfect value requires the normal distribution to collapse to a constant $c$ and all anomalies to exceed $c$.}
    \label{fig:Comparison_of_ideal_score_distributions_under_different_metrics}
\end{figure}

Regarding the second limitation, perfect anomaly detection fundamentally requires separating positive (anomalous) and negative (normal) distributions of anomaly scores completely. Both \citet{lunardi2025contrastiverepresentationmodelinganomaly} and \citet{jiang2023ScoreDistributionDiscrimination} emphasize such similar concepts. 
Without explicitly measuring score-distribution separability, one cannot determine whether a high evaluation score arises from a genuine separation between normal and anomalous samples or merely from a favorable ranking effect. Moreover, when the two score distributions overlap substantially, samples in the overlapping region become ambiguous, making reliable discrimination more difficult~\citep{Metzner2022DataAmbiguity}.
However, no existing metric directly quantifies such distributional separability. AUROC assesses score ranking~\citep{maurer2020ROCestimating}, while AUPR and AUPRO evaluate precision-recall trade-offs and region overlap, respectively. 

Regarding the third limitation, existing metrics largely bypass the candidate threshold selection problem. AUROC is threshold-independent because it integrates over all thresholds, ignoring the difficulty of selecting possible thresholds in practice. As shown in Fig.~\ref{fig:Comparison_of_ideal_score_distributions_under_different_metrics} (a--b), AUROC$=1.0$ can correspond to either a negligibly small or a wide feasible threshold margin, but a larger margin clearly provides better robustness in threshold selection. AUPRO partially addresses threshold selection by restricting the evaluation to a low-FPR range (e.g. FPR $\le$ 0.05). However, this range includes extremely low FPRs corresponding to very high thresholds, which can lead to an unacceptable increase in the False Negative Rate (FNR) or decrease in the recall. A detailed visual analysis is provided in Supplementary Material Section~S6.
Meanwhile, calculating FPR is a posterior process that requires access to anomaly labels, which is not possible in common unsupervised settings. Consequently, existing metrics do not capture model performance under realistic deployment conditions, where the candidate threshold range is necessarily derived from anomaly-free samples alone.

To address these limitations, we propose the \textbf{Bounded Area Under the Kolmogorov--Smirnov (Bounded AUKS)} curve. We derive AUKS from the Kolmogorov--Smirnov(KS) integral as a direct measure of separability between normal and anomalous scores distributions, equivalent to the 1-Wasserstein distance. 
We define the candidate threshold interval exclusively from normal samples, consistent with the unsupervised deployment setting in which anomalous labels are unavailable for threshold selection. The lower bound $\theta_{\min}=\mathrm{Median}+\alpha\,\mathrm{MAD}$ is motivated by robust statistics and filters out the low-threshold region that is dominated by high FPR yet provides limited discriminative information. The upper bound $\theta_{\max}$ defines the upper limit of the interval and excludes excessively high thresholds, since the segmentation corresponding to the thresholds near the maximum value can be overly conservative and the recall can be markedly reduced. We use Extreme Value Theory(EVT) to determine $\theta_{\max}$ at tail-risk level $q$ because this upper bound is governed by rare upper-tail of normal-scores. Bounded AUKS integrates the difference between TPR and FPR strictly over the interval defined by the bounded thresholds region.

The contributions of this paper are as follows:
\begin{itemize}
\item We propose \textbf{Bounded AUKS}, a bounded KS-integral metric that quantifies anomaly score distribution separability within a threshold range, addressing the limitations of ranking-based evaluation.
\item We introduce \textbf{anomaly-free threshold bounds} inferred solely from anomaly scores of normal test samples, including a robust lower bound and an EVT-based upper bound, to define a practically relevant evaluation interval without requirements of known anomalies.
\item We provide \textbf{numerical studies} on MVTec AD and VisA datasets, demonstrating that Bounded AUKS has a better performance in evaluating unsupervised anomaly detection models.
\end{itemize}

The remainder of this paper is organized as follows. Section~\ref{sec:related_work} reviews related work on anomaly detection evaluation metrics and threshold selection. Section~\ref{sec:method} presents the proposed Bounded AUKS and its properties. Section~\ref{sec3} describes the experimental settings and reports the main results. Section~\ref{sec:conclusion and discussion} concludes the paper.

\section{Related Work}\label{sec:related_work}



The choice of anomaly-detection evaluation metrics remains actively debated. \citet{fourure2021anomalydetectionartificiallyincrease} showed that non-standardized protocols and varying anomaly rates can substantially inflate F1 and AUPR, undermining comparisons across datasets. At the pixel level, \citet{Rafiei2023PixelLevel} found that precision--recall-based metrics better reflect relative localization performance under severe class imbalance. For imbalanced data, \citet{Ma2025iAPciteAUPIMO} recommended average precision across recall levels and the Dice similarity coefficient.

Recent work has also introduced specialized metrics. \citet{Zhang2023DeSTSeg} proposed Instance Average Precision (IAP), which defines region-level recall by treating a region as detected when at least half of its pixels are correctly identified. IAP is also used for threshold selection, while pixel-level precision at fixed recall (precision-at-\emph{k}\%) supports model comparison. Nevertheless, IAP remains a ranking-based precision--recall metric and relies on a heuristic 50\% overlap criterion. \citet{Bertoldo2024AUPIMO} proposed the Area Under the Per-Image Overlap (AUPIMO) curve, an AUROC-derived recall metric that validates models using only normal images, reducing bias toward previously observed anomalies. Its per-image recall supports both individual- and cross-image analyses. However, AUPIMO evaluates localization recall within prescribed lower and upper FPR bounds. This interval remains parameterized by FPR, an a posteriori quantity that depends on access to anomaly labels; like AUROC, AUPIMO primarily measures score ranking rather than the separability of normal- and anomalous-score distributions. Focusing instead on industrial operating points, \citet{Baitieva2025BeyondAcademic} introduced two image-level ROC points: Presorted Good at 2\% (PG2), defined as TNR at FNR $=0.02$, and Presorted Bad at 2\% (PB2), defined as TPR at FPR $=0.02$ (where good denotes normal and bad denotes anomalous). Because these are point estimates at fixed thresholds, they do not summarize performance across a range of operating conditions.

\citet{wilkinghoff2023f1ev} proposed the F1-expected value (F1-EV) score to quantify threshold-estimation difficulty and a bounded variant for realistic operating points. Its integration interval uses the lower bound $\mu-\alpha\sigma$ and the upper bound $\theta_{\text{opt}}+\alpha\sigma$, where $\mu$ and $\sigma$ are the mean and standard deviation of normal scores and $\theta_{\text{opt}}$ maximizes F1. Although this construction reduces the influence of outliers and correlates strongly with deployment performance, it assumes a Gaussian-like normal-score distribution and may therefore be unsuitable for heavy-tailed or skewed scores. Moreover, using the ground-truth $\theta_{\text{opt}}$ requires anomalous-sample information and introduces hindsight bias. The lower bound is also questionable in practice: thresholds below $\mu$ would classify approximately half of the normal samples as anomalous. Supplementary Material Section~S1 provides further discussion of threshold selection.

\section{Bounded AUKS Metric}\label{sec-meth}
\label{sec:method}

In this section, we formalize the proposed metric. We start by defining the Area Under the Kolmogorov--Smirnov curve (AUKS) as a directed separability measure derived from the gap between distributions of normal-scores and anomalous-scores. We then introduce \textbf{Bounded AUKS}, which refines this measure by restricting evaluation to a operational range $[\theta_{\text{Lower}}, \theta_{\text{Upper}}]$ derived exclusively from anomaly scores of normal samples.

\subsection{Score distributions and TPR--FPR}

Let $S$ be the continuous anomaly score output by the model. We denote the scores of normal samples as $S_{\text{norm}}$ and those of anomalous samples as $S_{\text{anom}}$. Denote by $F_{\text{norm}}(\theta)$ and $F_{\text{anom}}(\theta)$ the Cumulative Distribution Function(CDF) of $S$ for normal and anomalous samples, respectively. Using the decision rule “predict anomaly if $S > \theta$”, we have:
\[
\mathrm{TPR}
(\theta) = \mathbb{P}(S > \theta \mid \text{anom}) = 1 - F_{\text{anom}}(\theta), \quad
\mathrm{FPR}(\theta) = \mathbb{P}(S > \theta \mid \text{norm}) = 1 - F_{\text{norm}}(\theta).
\]
Consequently, the pointwise gap between the CDFs corresponds directly to the gap between TPR and FPR:
\begin{equation}
F_{\text{norm}}(\theta) - F_{\text{anom}}(\theta) = \mathrm{TPR}(\theta) - \mathrm{FPR}(\theta).
\label{eq:cdf_tpr_fpr_link}
\end{equation}

The standard Kolmogorov-Smirnov statistic~\citep{Massey01031951ks} evaluates distribution separability by measuring the maximum absolute difference between the CDFs:
\begin{equation}
\mathrm{KS \ statistic} = \max | F_{\text{norm}}(\theta) - F_{\text{anom}}(\theta) | = \max |\mathrm{TPR}(\theta) - \mathrm{FPR}(\theta)|.
\end{equation}

However, the absolute value is inappropriate for anomaly detection because it removes the direction of separation. Consequently, inverted score distributions can still produce a large KS value even when normal samples are assigned higher anomaly scores than anomalous ones. We therefore use the signed gap \( \mathrm{KS}(\theta) = \mathrm{TPR}(\theta) - \mathrm{FPR}(\theta) \), so that positive values indicate the desired ordering and negative values indicate inversion. 

\subsection{AUKS: integral separability and mathematical grounding}

We define the area under the KS curve (\textbf{AUKS}) as the integral of KS over the entire normalized score domain $[0,1]$:
\begin{equation}
\mathrm{AUKS}
= \int_{0}^{1} \big(F_{\text{norm}}(\theta) - F_{\text{anom}}(\theta)\big)\, d\theta
= \int_{0}^{1} \big(\mathrm{TPR}(\theta) - \mathrm{FPR}(\theta)\big)\, d\theta.
\label{eq:auks_def}
\end{equation}

In an effective anomaly detection setting, models should assign consistently higher scores to anomalous samples, establishing the ideal stochastic ordering $F_{\text{norm}}(\theta) \ge F_{\text{anom}}(\theta)$ for any applicable threshold $\theta$. Therefore, we purposely omit the absolute value inside the integral. A larger AUKS thus mathematically guarantees that the anomaly score distribution is structurally separated to the right of the normal distribution, providing a robust global summary of separability.

AUKS also has an optimal-transport interpretation. For one-dimensional distributions on $[0,1]$, the 1-Wasserstein distance is $W_1(F_{\text{norm}},F_{\text{anom}})=\int_0^1|F_{\text{norm}}(\theta)-F_{\text{anom}}(\theta)|\,d\theta$~\citep{Panaretos2019StatisticalWasserstein,Vallender1974WassersteinDistance}. Under the stochastic ordering assumed above, the absolute value can be removed, and AUKS therefore equals $W_1$. When this ordering is violated, the signed AUKS instead penalizes score inversion and need not equal the Wasserstein distance.

\subsection{Bounded AUKS: Anomaly-free threshold bounds}

While AUKS measures global separability, practical deployment requires evaluating models only within a reliable $\theta$ range. We propose \textbf{Bounded AUKS}, which restricts the integration to a data-driven interval $[\theta_{\text{Lower}}, \theta_{\text{Upper}}]$ inferred solely from the data of normal samples.

\begin{figure}[!ht]
    \centering
    \includegraphics[width= \linewidth]{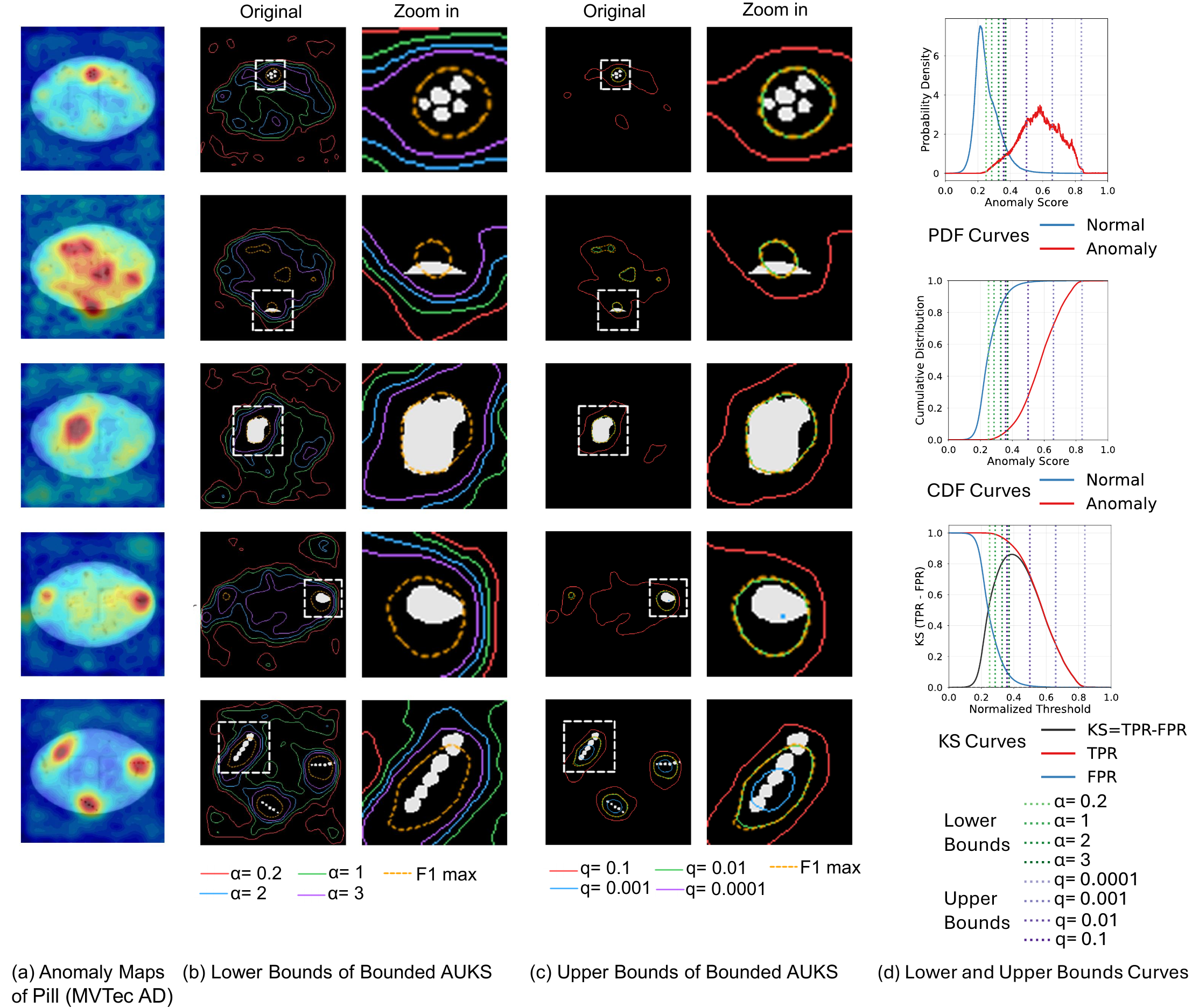}
    \caption{Impact of bounding parameters on segmentation bounds and metric curves for the MVTec AD Pill. (a) Corresponding anomaly map. (b) Lower bounds derived using different $\alpha$. (c) Upper bounds configured using varying EVT risk levels $q$. (d) Alignment of these bounds along the corresponding curves.}
    \label{fig:muti_para_bounds}
\end{figure}

The Bounded AUKS is formulated as the average signed separability over this defined interval:
\begin{equation}
\mathrm{Bounded\,AUKS}
= \frac{1}{\theta_{\text{Upper}} - \theta_{\text{Lower}}} \int_{\theta_{\text{Lower}}}^{\theta_{\text{Upper}}} \big(\mathrm{TPR}(\theta) - \mathrm{FPR}(\theta)\big)\, d\theta.
\label{eq:bounded_auks_def}
\end{equation}
The proposed bounds are as follows:
\begin{equation}
\begin{aligned}
\theta_{\text{Lower}} &= \mathrm{Median}(S_{\text{norm}}) + \alpha \cdot \mathrm{MAD}(S_{\text{norm}}) , \quad
\theta_{\text{Upper}} &= \text{EVT}(S_{\text{norm}}, q)
\end{aligned}
\label{eq:bounds_summary}
\end{equation}
where $\alpha$ and $q$ are tunable hyperparameters,  $\mathrm{MAD}$ denote the median absolute deviation. We provide the detailed justification for each bound below.

Figure~\ref{fig:muti_para_bounds} illustrates the effect of different parameters for Bounded AUKS on the resulting segmentation and corresponding integration bounds, using the Pill category of MVTec AD dataset as an example. The multi-panel figure shows how variations in the bounds influence the spatial segmentation and map onto the Probability Density Function (PDF), CDF, and KS curves. For the upper bound, regions corresponding to excessively high thresholds are excluded, thereby preventing a substantial reduction in TPR. We observe that our chosen setting ($q=0.01$) produces an appropriate segmentation and is reasonably close to the theoretically F1-max optimal point. If the allowable risk $q$ is set too large (e.g., $q=0.1$), the resulting threshold becomes excessively low, leading to a loose segmentation with many false positives and potentially causing the upper bound to approach or even fall below the lower bound. Conversely, if $q$ is excessively small (e.g., $q=0.0001$), the resulting threshold becomes overly stringent, causing the segmentation to contract to such an extent that the anomaly may fail to be detected altogether.
In Figure~\ref{fig:muti_para_bounds} (d), we observe that an excessively restrictive $q$ provokes a steep collapse in TPR (falling below $0.7$), whereas optimal lower bounds successfully sidestep the high-FPR regime (FPR remaining below about $0.15$) while retaining a broad and informative integration interval.
Regarding the lower bound, which is governed by the multiplier $\alpha$, its primary function is to filter out the high FPR region, while guaranteeing the residual integration interval under the KS curve remains sufficiently wide to capture the separability of distributions.

\subsubsection{Upper Bound: EVT-Based Bound and Recall Preservation}
The upper bound $\theta_{\text{Upper}}$ defines a principled upper limit for the threshold interval used in Bounded AUKS. Because anomalous labels are unavailable in the unsupervised setting, we avoid threshold selection strategies that depend on FPR-indexed evaluation protocols, such as those used in AUPRO assessment. Instead, the selected threshold should be determined solely by the statistical behavior of normal data, which is also the reason that our analysis is only based on normal-scores. Since the threshold is primarily determined by the extreme upper tail of the normal-score distribution, where samples are scarce, direct estimation via empirical quantiles can be unstable in the high-confidence regime. We therefore employ EVT to model the tail, rather than imposing a strong parametric assumption on the entire distribution. Moreover, EVT enables us to specify a risk level $q$ that defines a nominal false alarm probability and thereby yielding a principled threshold. When the tail model is well specified and the sample size is sufficiently large, the resulting FPR approaches the prescribed risk level $q$.

\textbf{Normal-score distribution tail modeling.}
Let $S_{\text{norm}}$ denote normal-scores and select a high initial threshold $u$ (e.g., we use the $90$th percentile). We then focus on the excesses of normal-scores above $u$, defined as
$ Y = S_{\text{norm}} - u \,\big|\, S_{\text{norm}} > u $.

Under the Peaks-over-Threshold (POT) framework, these excesses can be modeled by a Generalized Pareto Distribution (GPD) with shape parameter $\xi$ and scale parameter $\sigma$, whose cumulative distribution function is given by
\begin{equation}
\mathbb{P}(Y \le y) = 1 - \left(1+\xi y/\sigma\right)^{-1/\xi}, \qquad y>0,
\label{eq:gpd_cdf}
\end{equation}
where $\xi>0$ corresponds to a heavy tail (unbounded support). Let $\zeta_u=\mathbb{P}(S_{\text{norm}}>u)$ be the exceedance probability. Supplementary Material Section~S3, Table~S3 summarizes the key parameters used to instantiate this tail model and the resulting upper bound.

\textbf{EVT-based upper bound definition.}
Let $q \in (0,\zeta_u]$ denote a target tail risk level, which serves as a nominal false alarm probability under the EVT tail model. Under the POT-GPD approximation, the upper bound $\theta_{\text{Upper}}$ is defined as:
$ \mathbb{P}(S_{\text{norm}} > \theta_{\text{Upper}}) = q $,
Using the standard POT tail approximation
\begin{equation}
\mathbb{P}(S_{\text{norm}} > \theta)
=
\zeta_u \left(1+\xi(\theta-u)/\sigma\right)^{-1/\xi},
\qquad \theta \ge u,
\label{eq:pot_tail_survival}
\end{equation}
we obtain
\begin{equation}
\theta_{\text{Upper}}(q)
=
u+\frac{\sigma}{\xi}\left[\left(\frac{q}{\zeta_u}\right)^{-\xi}-1\right].
\label{eq:theta_upper_closed_form}
\end{equation}
In the small-risk regime $q \ll \zeta_u$,
\begin{equation}
\theta_{\text{Upper}}(q)
\sim
\frac{\sigma \zeta_u^{\xi}}{\xi} q^{-\xi},
\qquad q \to 0,
\label{eq:theta_upper_asymptotic}
\end{equation}
and hence $\theta_{\text{Upper}}(q)=\Theta(q^{-\xi})$. With $N$ normal samples and $N_u$ exceedances above $u$, we use $\hat{\zeta}_u = N_u/N$ together with fitted $(\hat{\xi},\hat{\sigma})$, yielding
\begin{equation}
\hat{\theta}_{\text{Upper}}(q)
=
u+\frac{\hat{\sigma}}{\hat{\xi}}
\left[
\left(\frac{qN}{N_u}\right)^{-\hat{\xi}}-1
\right].
\label{eq:theta_upper_plugin}
\end{equation}

\textbf{Recall--risk trade-off and practical choice of $q=0.01$.}
Decreasing $q$ makes the EVT upper bound increasingly conservative. If the bound moves beyond most anomalous scores, recall drops sharply, producing what we term the \emph{recall cliff}. Parameterized simulations and empirical recall--risk curves on MVTec AD Pill and VisA pcb1 consistently exhibit this low-$q$ collapse. Together with Figure~\ref{fig:muti_para_bounds}, these results motivate $q=0.01$ as a stable setting in our experiments. This value avoids the collapse associated with smaller values without producing the loose upper bound observed at $q=0.1$. The complete derivation, simulation design, and all simulated and empirical curves are provided in Supplementary Material Section~S3.1.

\subsubsection{Lower Bound: Robust Statistics and Parameter Mapping}

We adopt a robust formulation: $\theta_{\text{Lower}} = \mathrm{Median} + \alpha \cdot \mathrm{MAD}$.
We rely on Median and MAD rather than Mean and Standard Deviation for two key reasons grounded in robust statistics. First, anomaly scores typically follow a heavy-tailed, right-skewed distribution where the mean is heavily pulled towards the tail. Our median-based formulation inherently adapts to this skewness by targeting the right side of the central mass, effectively skipping the uninformative bulk. Second, Median and MAD provide superior resistance to outliers. Unlike the mean, which can be drastically distorted by even a single extreme prediction error (a 0\% breakdown point), the median and MAD remain stable as long as less than half of the scores are extreme (a 50\% breakdown point). This guarantees that our lower bound will not be artificially inflated by a few occasional noisy normal samples.

\textbf{Proposed Mapping of $\alpha$ to FPR.}
To mathematically map $\alpha$ to the percentile $p$, which represents the percentage of anomaly scores of normal samples sorted from smallest to largest (i.e., FPR $= 1 - p$), we must approximate the MAD.

We model the normal-score distribution using the \textbf{Fréchet distribution} ($F(x) = \exp(-(x/s)^{-\gamma})$)~\citep{coles2001introduction}. We adopt this specific distribution because, in EVT, the Fréchet distribution is the standard macroscopic model for strictly positive data with heavy, polynomial-decay tails. Thus, while $\theta_{\text{Upper}}$ uses POT and GPD locally to rigorously constrain the extreme upper excesses, the Fréchet distribution serves as its natural global counterpart to characterize the overall right-skewed behavior of anomaly scores.

Since normal-score distributions are bounded by zero on the left and highly skewed to the right, negative deviations from the median are strictly compressed. As a result, the absolute deviations are overwhelmingly dominated by the upper half of the data. Therefore, we safely approximate MAD as the distance between the upper quartile ($75^{\text{th}}$ percentile, $x_{0.75}$) and the Median ($50^{\text{th}}$ percentile, $x_{0.5}$): $\mathrm{MAD} = x_{0.75} - x_{0.5}$. Under the Fréchet assumption, the ratio $R = x_{0.75}/x_{0.5}$ elegantly cancels out the scale parameter $s$ and depends solely on the shape parameter $\gamma$ (where lower $\gamma$ signifies heavier right tails):
\[
R = \left( \frac{\ln(1/0.75)}{\ln(1/0.5)} \right)^{-1/\gamma} = \left( \frac{\ln 4 - \ln 3}{\ln 2} \right)^{-1/\gamma}.
\]
Substituting $\theta_{\text{Lower}} = x_{0.5} [1 + \alpha(R-1)]$ into the CDF yields the explicit mapping:
\begin{equation}
\label{eq:frechet_alpha}
p(\alpha) = F(\theta_{\text{Lower}}) = \exp\left( -\ln 2 \cdot \left[ 1 + \alpha \cdot \left( \left( \frac{\ln 4 - \ln 3}{\ln 2} \right)^{-1/\gamma} - 1 \right) \right]^{-\gamma} \right).
\end{equation}

Table~\ref{tab:alpha_percentile_frechet} presents the resulting FPR for $\alpha$, serving as a reference for parameter selection.

\begin{table}[h]
\centering
\caption{Mapping $\alpha$ to FPR under Fréchet Distribution.}
\label{tab:alpha_percentile_frechet}
\fontsize{9pt}{11pt}\selectfont
\renewcommand{\arraystretch}{0.8}
\begin{tabular*}{\linewidth}{@{\extracolsep{\fill}}lcccccc@{}}
\toprule
\textbf{Tail shape} & \multicolumn{6}{c}{\textbf{$\boldsymbol{\alpha}$}} \\
\cmidrule(lr){2-7}
$\boldsymbol{\gamma}$ & 0.2 & 1.0 & 2.0 & 3.0 & 4.0 & 5.0 \\
\midrule
4 (Light) & 37.8\% & 25.0\% & 10.9\% & 5.3\% & 2.8\% & 1.6\% \\
2 (Typical) & 37.0\% & 25.0\% & 13.5\% & 9.2\% & 5.1\% & 3.5\% \\
1 (Extreme) & 36.6\% & 25.0\% & 15.6\% & 10.8\% & 8.0\% & 6.2\% \\
\bottomrule
\end{tabular*}
\end{table}

\subsection{Properties}

\textbf{Range and Interpretation.}
Bounded AUKS values range from -1 to 1. A value of 1 indicates perfect separation within the operational window. A value of 0 implies random performance ($\mathrm{TPR} = \mathrm{FPR}$), and negative values indicate systematic error where the model assigns higher scores to normal samples than to anomalies.

\medskip\noindent
\textbf{Compactness and Ideal Separation.}
While AUROC reaches 1.0 for any perfectly separated ranking regardless of distribution spread, AUKS and Bounded AUKS impose stricter geometrical constraints (see Figure~\ref{fig:Comparison_of_ideal_score_distributions_under_different_metrics}). 
\textbf{AUKS@100\% = 1.0} (Fig.~\ref{fig:Comparison_of_ideal_score_distributions_under_different_metrics}(c)) requires extreme polarization: all normal-scores must be exactly 0 and all anomalous-scores exactly 1, which is often theoretically unattainable. In contrast, \textbf{Bounded AUKS = 1.0} (Fig.~\ref{fig:Comparison_of_ideal_score_distributions_under_different_metrics}(d)) is asymptotically achievable under a specific condition: the normal-score distribution must collapse to a single constant $c$ (minimizing the intra-class variance), and all anomalous-scores must exceed $c$. Since we utilize upper and lower bounds to estimate the threshold, the metric inherently incorporates threshold selection information. In (d), our estimated threshold corresponds to $c$ (or is slightly larger than $c$).
    
It is worth noting that achieving a perfect Bounded AUKS value only requires anomalous samples to have scores exceeding $c$, which is consistent with the nature of anomalies, namely their diversity and varying levels of detection difficulty. Therefore, it's unrealistic to expect all the scores in the anomalous samples to be concentrated at the perfect score of 1. Similarly, suppressing the scores of all normal samples to 0 is also extremely difficult. Instead, we believe that normal patterns should be compactly clustered and a high Bounded AUKS value only requires the normal-scores to be tightly concentrated.

\section{Results}\label{sec3}
In this section, we present empirical results on two industrial anomaly detection benchmarks and compare Bounded AUKS with widely used evaluation metrics. After introducing the experimental settings in Section~4.1, we present quantitative comparisons, discuss representative case studies, and summarize the main findings.

\subsection{Experimental Settings}
We conduct experiments on two standard industrial benchmarks: MVTec AD and VisA. MVTec AD covers 15 categories (5 textures and 10 objects) with diverse anomaly types on high‑resolution images. Detailed per‑category statistics are provided in Supplementary Material Section~S4, Table~S4. VisA includes 12 object categories, over 9{,}000 images, and 19 anomaly types under real‑world manufacturing conditions. We follow the default VisA one‑class split used in our implementation, with per‑category counts summarized in Supplementary Material Section~S4, Table~S5.

The comparison covers representative unsupervised anomaly detection models: PaDiM \citep{Defard2021PaDiM}, PatchCore \citep{Roth2022PatchCore}, STFPM \citep{Wang2021STFPM}, Reverse Distillation \citep{Deng2022ReverseDistillation}, FastFlow \citep{Yu2021FastFlow}, EfficientAD-S/M \citep{Batzner2024EfficientAD}, SuperSimpleNet \citep{Rolih2025SuperSimpleNet}, UniNet \citep{Wei2025UniNet}, and Dinomaly \citep{Guo2025Dinomaly}. These models span density estimation, memory-bank retrieval, student--teacher distillation, normalizing flows, and reconstruction-based paradigms. A concise overview of the evaluated counterparts is given in Supplementary Material Section~S5, Table~S6. Implementations use Anomalib defaults \citep{Akcay2022Anomalib}. For each model and category, we compute pixel-level AUROC, AUPRO@5\%, AUPRO@30\%, AUKS@100\%, and Bounded AUKS with $\alpha=2$ and $q=0.01$.

\begin{figure}[!t]
    \centering
    \includegraphics[width=\linewidth]{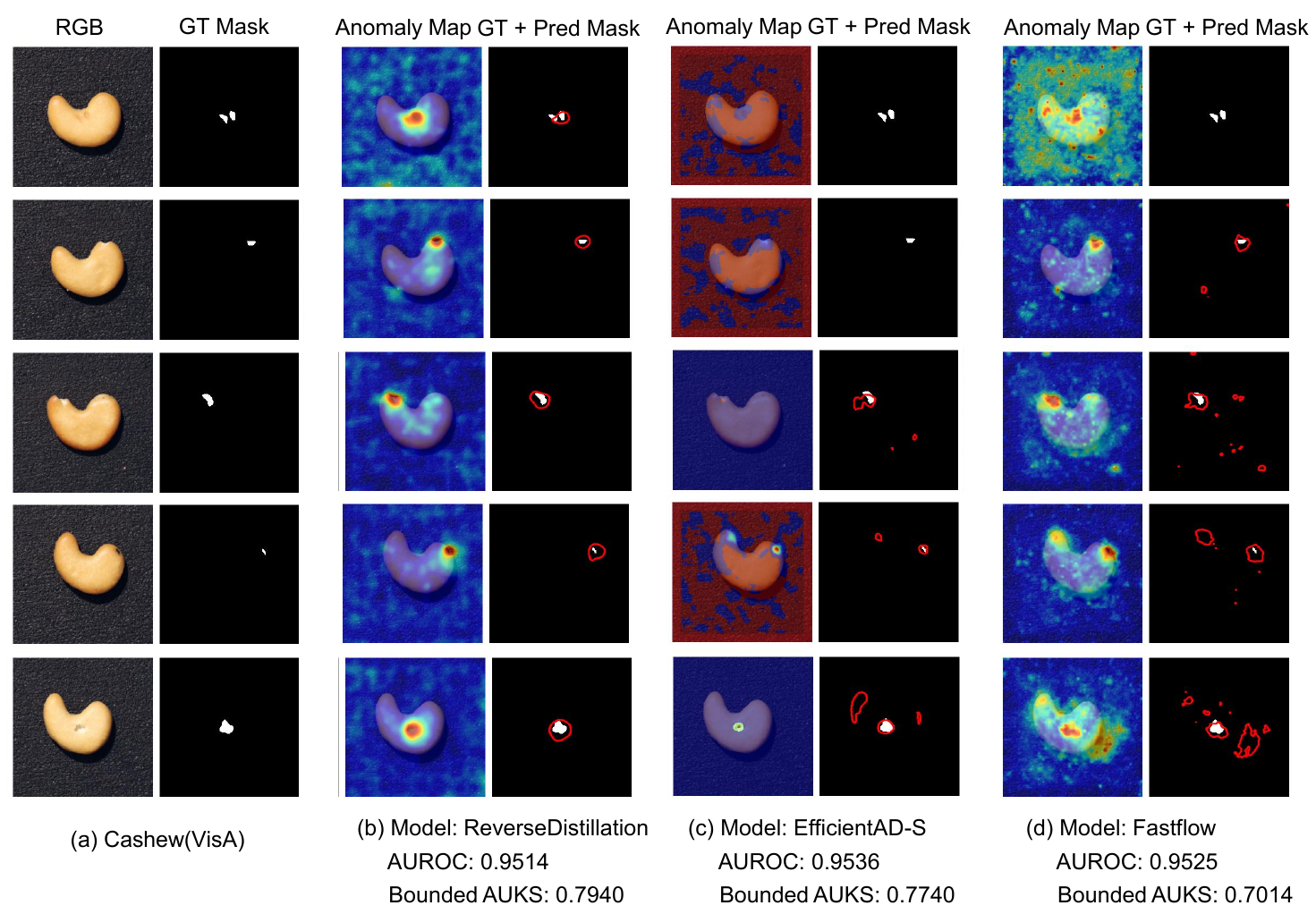}
    \caption{Case comparison on VisA (Cashew). (a) test images and ground truth masks. (b)--(d) Anomaly heatmaps and predicted mask segmentations at the F1-max optimal point overlaid on the ground-truth mask for (b) Reverse Distillation, (c) EfficientAD-S, and (d) FastFlow.}
    \label{fig:case_compare_models}
\end{figure}

\subsection{Cases Comparison Results}

\begin{figure*}[!ht]
    \centering
    \includegraphics[width=\textwidth]{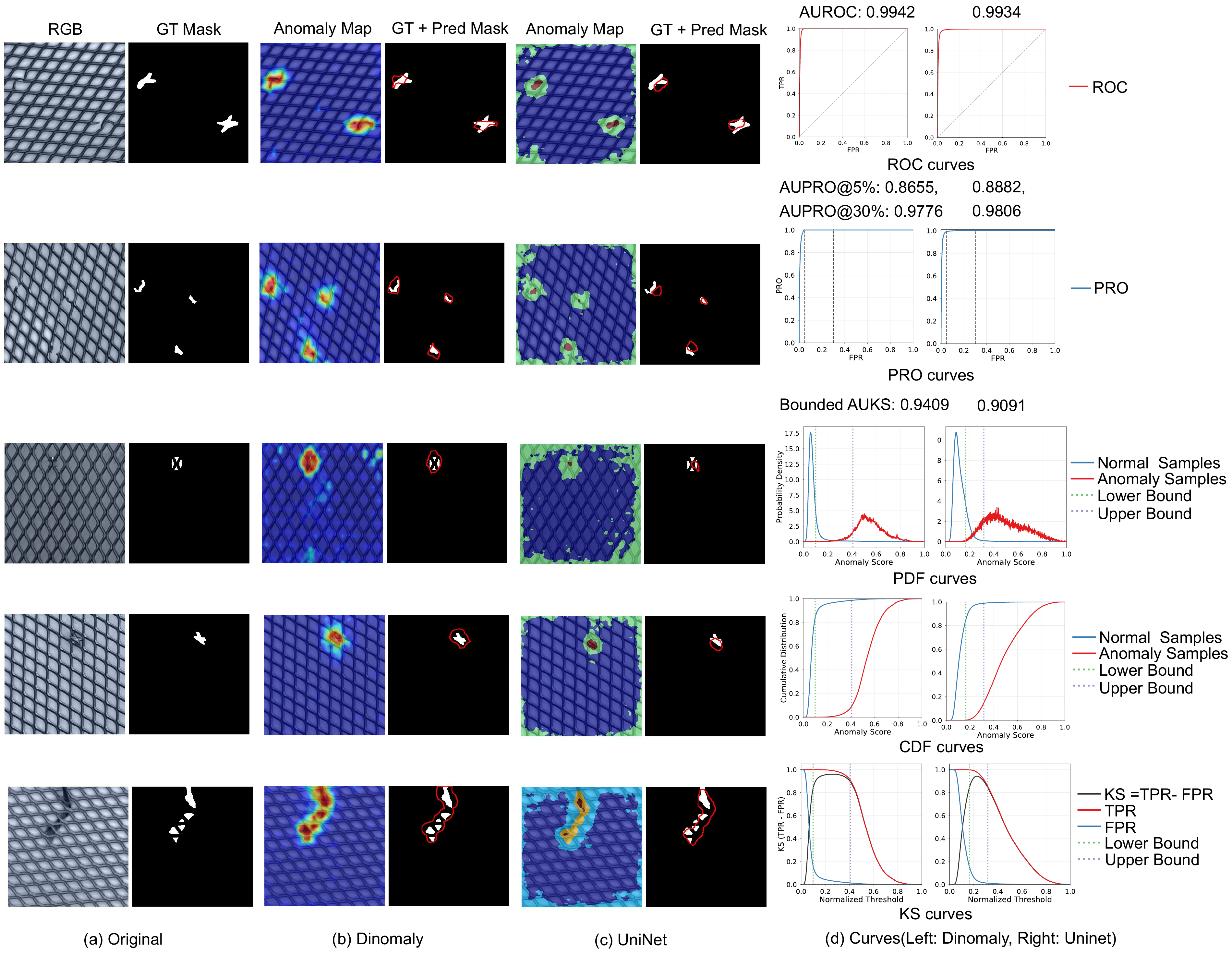}
    \caption{Comparison between Dinomaly and UniNet on the grid category of MVTec AD dataset. (a) Original RGB test images and ground truth masks. (b) (c) Anomaly maps and predicted mask segmentations overlaid on the ground truth mask for (b) Dinomaly and (c) UniNet. (d) Corresponding ROC, PRO, PDF, CDF, and KS curves for Dinomaly (left) and UniNet (right). The segmentation threshold for each model is set at the F1-max optimal point.
    }
    \label{fig:images_and_curves_compare}
\end{figure*}

Figure~\ref{fig:case_compare_models} shows a representative VisA Cashew case. AUROC ranks EfficientAD-S above FastFlow and Reverse Distillation with only marginal differences: 0.9536 for EfficientAD-S, 0.9525 for FastFlow, and 0.9514 for Reverse Distillation. Visual masks show the opposite ordering. Reverse Distillation produces the most concentrated heatmaps over the anomaly and the cleanest segmentation. EfficientAD-S shows occasional over- and under-segmentation, while FastFlow incorrectly flags large normal background regions as anomalous. Bounded AUKS produces the visually consistent order with scores 0.7940, 0.7740, and 0.7014 for Reverse Distillation, EfficientAD-S, and FastFlow.

Figure~\ref{fig:images_and_curves_compare} compares Dinomaly and UniNet on the MVTec AD grid category. Dinomaly produces cleaner anomaly maps and more accurate masks by suppressing background responses while preserving true anomalies, yet their AUROC values are nearly identical (0.9942 versus 0.9934), and both AUPRO@5\% and AUPRO@30\% incorrectly favor UniNet. Bounded AUKS instead assigns 0.9409 to Dinomaly and 0.9091 to UniNet, consistent with the visual evidence. The right-side PDF, CDF, and KS curves explain this result: Dinomaly achieves greater separation between normal- and anomalous-score distributions, whereas UniNet shows more overlap in the operative threshold region. Thus, Bounded AUKS evaluates whether score distributions support reliable localization rather than ranking alone.

The visual comparison of the threshold bounds used by Bounded AUKS and AUPRO is provided in Supplementary Material Section~S6.

\subsection{Metrics Performance Summary}

\begin{table*}[!t]
    \centering
    \caption{Overall average ranks and pixel-level metric values across all categories on MVTec AD and VisA.}
    \label{tab:overall_metric_ranks}
    \fontsize{9pt}{11pt}\selectfont
    \setlength{\tabcolsep}{4pt}
    \makebox[\textwidth][c]{%
    \begin{tabular}{lccccc}
        \toprule
        \multicolumn{6}{c}{\textbf{MVTec AD}} \\
        \midrule
        \textbf{Model} & \textbf{AUROC} & \textbf{AUPRO@5\%} & \textbf{AUPRO@30\%} & \textbf{AUKS@100\%} & \textbf{Bounded AUKS} \\
        \midrule
        Dinomaly & 1 (0.9845) & 1 (0.7924) & 1 (0.9535) & 3 (0.4059) & 1 (0.8329) \\
        PatchCore & 2 (0.9801) & 6 (0.6885) & 3 (0.9298) & 6 (0.3525) & 2 (0.7964) \\
        Reverse Distillation & 3 (0.9739) & 3 (0.7152) & 2 (0.9304) & 4 (0.3828) & 5 (0.7688) \\
        FastFlow & 4 (0.9738) & 5 (0.6941) & 4 (0.9138) & 2 (0.4120) & 7 (0.7574) \\
        UniNet & 5 (0.9721) & 7 (0.6812) & 6 (0.9022) & 5 (0.3555) & 4 (0.7693) \\
        SuperSimpleNet & 6 (0.9645) & 9 (0.6062) & 9 (0.8670) & 1 (0.4426) & 6 (0.7673) \\
        PaDiM & 7 (0.9632) & 10 (0.5816) & 7 (0.8974) & 7 (0.2342) & 9 (0.6371) \\
        EfficientAD-M & 8 (0.9579) & 2 (0.7392) & 5 (0.9027) & 8 (0.1811) & 3 (0.7708) \\
        EfficientAD-S & 9 (0.9497) & 4 (0.7118) & 8 (0.8790) & 10 (0.1478) & 8 (0.7465) \\
        STFPM & 10 (0.9473) & 8 (0.6224) & 10 (0.8647) & 9 (0.1549) & 10 (0.6345) \\
        \bottomrule
    \end{tabular}
    }

    \vspace{0.8em}

    \makebox[\textwidth][c]{%
    \begin{tabular}{lccccc}
        \toprule
        \multicolumn{6}{c}{\textbf{VisA}} \\
        \midrule
        \textbf{Model} & \textbf{AUROC} & \textbf{AUPRO@5\%} & \textbf{AUPRO@30\%} & \textbf{AUKS@100\%} & \textbf{Bounded AUKS} \\
        \midrule
        Dinomaly & 1 (0.9885) & 1 (0.8098) & 1 (0.9543) & 4 (0.3575) & 1 (0.8718) \\
        Reverse Distillation & 2 (0.9854) & 2 (0.7034) & 2 (0.9112) & 3 (0.3735) & 2 (0.8636) \\
        EfficientAD-M & 3 (0.9834) & 3 (0.6484) & 3 (0.8828) & 8 (0.1729) & 3 (0.8420) \\
        PatchCore & 4 (0.9804) & 8 (0.5708) & 4 (0.8616) & 6 (0.3177) & 4 (0.8036) \\
        EfficientAD-S & 5 (0.9726) & 7 (0.5841) & 6 (0.8240) & 9 (0.1363) & 6 (0.7925) \\
        FastFlow & 6 (0.9705) & 6 (0.5890) & 5 (0.8452) & 2 (0.3801) & 8 (0.7820) \\
        UniNet & 7 (0.9652) & 5 (0.5896) & 7 (0.8221) & 5 (0.3352) & 7 (0.7873) \\
        PaDiM & 8 (0.9618) & 10 (0.3916) & 9 (0.7509) & 7 (0.2100) & 9 (0.6603) \\
        SuperSimpleNet & 9 (0.9604) & 4 (0.6092) & 8 (0.8149) & 1 (0.4018) & 5 (0.8004) \\
        STFPM & 10 (0.8843) & 9 (0.4387) & 10 (0.6652) & 10 (0.1346) & 10 (0.5433) \\
        \bottomrule
    \end{tabular}
    }
\end{table*}

\begin{figure*}[!t]
    \centering
    \includegraphics[width=0.9\textwidth]{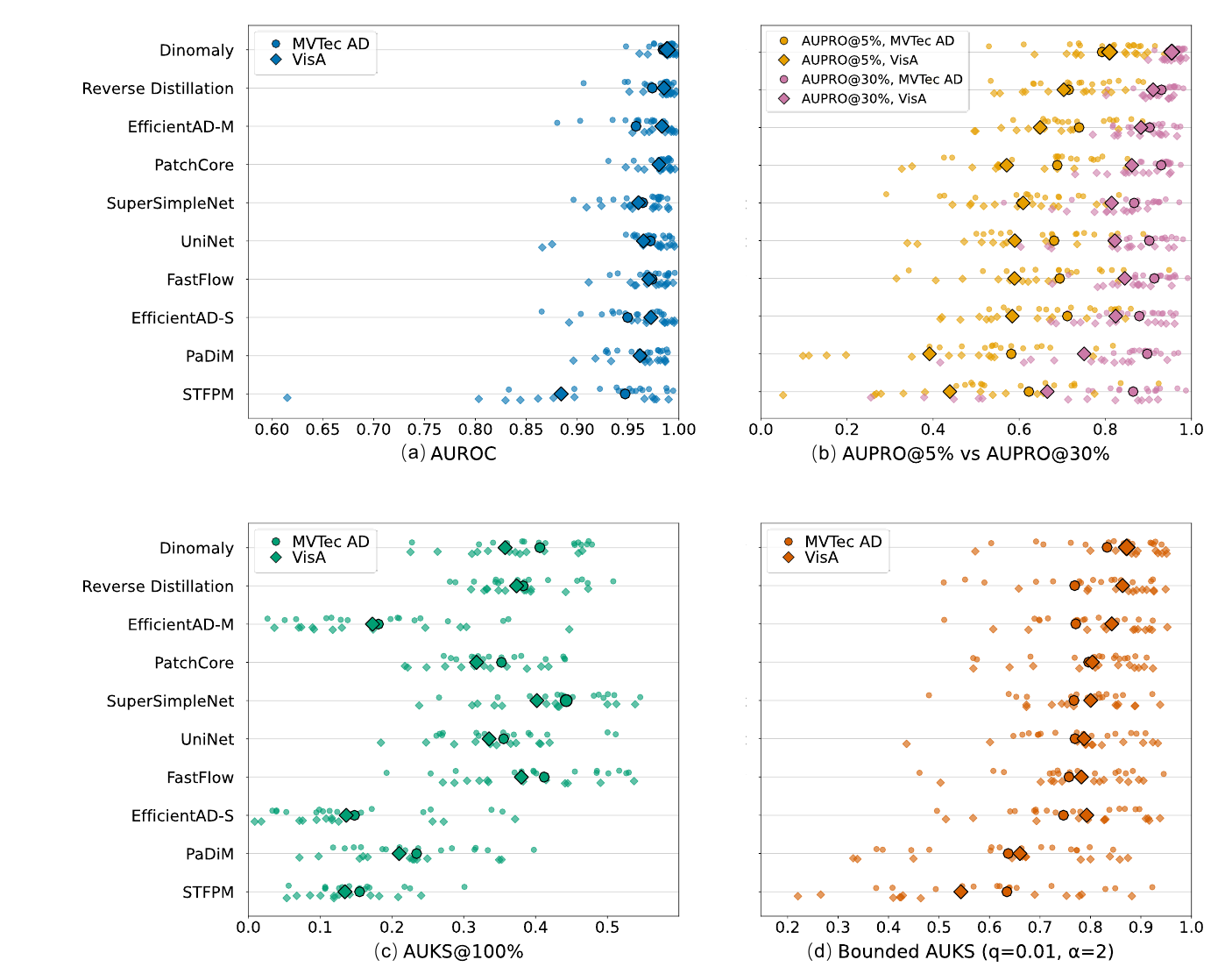}
    \caption{Pixel-level metric scatter summaries across categories on MVTec AD and VisA. (a) AUROC. (b) AUPRO@5\% vs AUPRO@30\%. (c) AUKS@100\%. (d) Bounded AUKS($q=0.01$, $\alpha=2$). Points denote per-category scores and mean markers lie on each model baseline; marker shape distinguishes datasets.}
    \label{fig:pixel_scatters}
\end{figure*}

Table~\ref{tab:overall_metric_ranks} provides the summary of model rankings and mean scores across all categories on MVTec AD and VisA. Each entry is reported as average rank followed by the mean score in parentheses. Figure~\ref{fig:pixel_scatters} complements this quantitative result with per-category scatter plots, so that both dataset-level averages and cross-category variations can be jointly analyzed.

We summarize the main findings as follows.
First, AUROC is near ceiling for many methods and the per-category scatters cluster tightly with little difference.
Second, restricting the FPR range to different limits produces different evaluation results. Comparing the AUPRO operations at range [0, 0.05] and [0, 0.30] reveals that imposing a tighter restriction on the FPR thresholds results in lower AUPRO values across all models, demonstrating that the metric is highly dependent on the chosen FPR thresholds.
Third, as shown in Table~\ref{tab:overall_metric_ranks}, the model rankings produced by AUKS@100\% exhibit significant discrepancies compared to other metrics, and its evaluation does not align well with actual qualitative performance. The primary reason is that AUKS@100\% integrates over the entire threshold range, incorporating many undesirable intervals i.e., regions where both TPR and FPR are simultaneously high, or simultaneously low. These regions are largely uninformative and can therefore distort the evaluation, resulting in biased performance estimates.

In particular, Bounded AUKS confines the integration to a strictly bounded interval, which effectively functions as an decision-threshold selection mechanism. Meanwhile, our Bounded AUKS is a stricter metric for quantifying the separability between normal- and anomalous-score distributions; its values span a wider range, distinctly highlighting the performance gaps between these models and thereby ranking the indeed better detection models.

\section{Conclusion}
\label{sec:conclusion and discussion}

Bounded AUKS evaluates anomaly localization through signed score-distribution separability over an anomaly-free threshold range. Unlike AUROC, fixed FPR windows, and post-hoc optimal thresholds, it directly assesses robust thresholding. Experiments on MVTec AD and VisA show that it reveals differences hidden by saturated metrics and explains qualitative localization behavior. The same principle can extend to other scored anomaly-detection settings and potentially serve as a training objective.

In application, Bounded AUKS supports model selection, threshold-range design, and deployment auditing. For detectors with similar conventional scores, it favors models with persistent score separation across a usable interval. Normal-score-derived bounds can be updated without anomalous samples, while $q$ links the interval to a nominal false-alarm tolerance. The metric thus connects offline comparison to practical threshold decisions.

\section*{Data Availability Statement}
\begingroup
\fontsize{10pt}{12pt}\selectfont
The MVTec AD dataset is available from \url{https://www.mvtec.com/research-teaching/datasets/mvtec-ad}; the VisA dataset is available through \url{https://github.com/amazon-science/spot-diff}. The dataset providers' terms of use apply.
\par\endgroup

\section*{Supplementary Material}
\begingroup
\fontsize{10pt}{12pt}\selectfont
The Supplementary Material provides supporting analyses and additional experimental details.
\par\endgroup


\begingroup
\footnotesize
\bibliography{bibliography.bib}
\endgroup

\end{document}